\documentclass[10pt,twocolumn,letterpaper]{article}

\usepackage{cvpr}              
\usepackage{multirow}
\usepackage{indentfirst}
\usepackage{caption}
\usepackage[dvipsnames]{xcolor}

\definecolor{cvprblue}{rgb}{0.21,0.49,0.74}
\usepackage[pagebackref,breaklinks,colorlinks,citecolor=cvprblue]{hyperref}

\def\paperID{12333} 
\def\confName{CVPR}
\def\confYear{2024}

\title{Implicit Discriminative Knowledge Learning for Visible-Infrared Person Re-Identification}

\author{Kaijie Ren, Lei Zhang*\\
	School of Microelectronics and Communication Engineering, Chongqing University, China\\
	{\tt\small kaijieren@cqu.edu.cn, leizhang@cqu.edu.cn}
}

\begin{document}
\maketitle
\begin{abstract}
	Visible-Infrared Person Re-identification (VI-ReID) is a challenging cross-modal pedestrian retrieval task, due to significant intra-class variations and cross-modal discrepancies among different cameras.
	Existing works mainly focus on embedding images of different modalities into a unified space to mine modality-shared features. They only seek distinctive information within these shared features, while ignoring the identity-aware useful information that is implicit in the modality-specific features.
	To address this issue, we propose a novel Implicit Discriminative Knowledge Learning (IDKL) network to uncover and leverage the implicit discriminative information contained within the modality-specific.
	First, we extract modality-specific and modality-shared features using a novel dual-stream network. Then, the modality-specific features undergo purification to reduce their modality style discrepancies while preserving identity-aware discriminative knowledge. Subsequently, this kind of implicit knowledge is distilled into the modality-shared feature to enhance its distinctiveness.
	Finally, an alignment loss is proposed to minimize modality discrepancy on enhanced modality-shared features. Extensive experiments on multiple public datasets demonstrate the superiority of IDKL network over the state-of-the-art methods. 
	Code is available at \url{https://github.com/1KK077/IDKL}.
\end{abstract}  
\section{Introduction}
\begin{figure}[t]
	\begin{center}
		\includegraphics[width=0.95\linewidth]{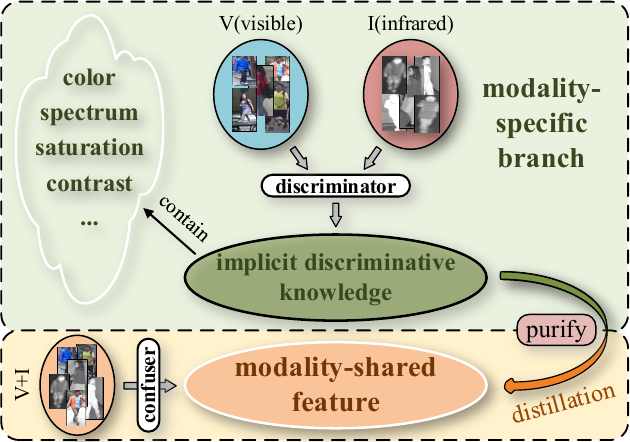}
	\end{center}
	\captionsetup{font=small}
	\caption{ Previous methods focused on seeking discriminative information within modality-shared features, overlooking the fact that there are discriminative clues implicit in modality-specific features. It is worth considering utilization of the implicit discriminative information to enhance shared invariant feature.}
	\label{fig:one}
	\vspace{5pt} 
\end{figure}
Visible-Infrared Person Re-Identification (VI-ReID) aims to match pedestrian images across multiple non-overlapping camera views and different modalities. Advanced surveillance systems today are capable of automatically switching from visible to infrared mode at nightfall, ensuring an ample supply of trainable data. However, the unique modality of infrared images create significant domain discrepancies and a more complex environment. This makes previous single-modality methods, which are based solely on visible images, less efficient for VI-ReID tasks.
Recently, numerous advanced methods have emerged in the field of VI-ReID. In conclusion, these approaches can generally be categorized into two types: the first aims to learn modality-shared features directly from raw modal data, and the second seeks to incorporate additional modal information to refine the feature space or bridge the modal gap, thereby facilitating the search for modality-shared features. 
Raw modal data based method \cite{zhu2020hetero, 9276429, hao2019hsme, 9711447, 9577446, 9710774, 9577837, li2022counterfactual} is to embed different modality images into the same space and align them on feature-level so that model learn modality invariant feature directly. 
Despite original images based methods has achieved desirable results, they still remain a huge gap between different modalities in feature space. In order to bridge the gap between visible and infrared modalities and construct a continuous space for better learning modality-shared features, various methods based on introducing extra information \cite{8953262, zhang2022modality, 9880449, jiang2022cross, yang2020cross, wang2020cross, 9115075} have emerged constantly.

Although significant improvements have been made in current VI-ReID methods, these models inevitably discard some discriminative information that relies on modality-specific features, which is not fully exploited and utilized previously. These discriminative cues that exist in modality-specific features can be referred to as implicit discriminative information, as shown by the green area in \cref{fig:one}, such as color, grayscale spectrum, contrast, saturation, and so on.
Therefore, relying solely on modality-shared cues can limit the upper boundary of the discrimination ability of the feature representation. Utilizing implicit modality-specific characteristics effectively is essential to enhance the distinctiveness of modality-invariant features. However, we cannot directly use the modality-specific 
knowledge due to the modality discrepancies it contains. Instead, we need to reduce these discrepancies while preserving the identity-aware discriminative information inherent in the implicit knowledge. Meanwhile, traditional VI-ReID methods involving distillation, alignment, and mutual learning typically rely on logits \cite{9935815,9107428,zhang2018deep}. However, there is no involvement of the classifier during the testing phase, with matching performed only at the feature level. Therefore, conducting discriminative information distillation at the feature level is also essential.

To address the above limitations, 
in this paper, we propose an Implicit Discriminative Knowledge Learning (IDKL) framework that captures implicit invariant information from modality-specific features and distills it into modality-shared features to enhance their discriminative capability. 
We first extract modality-specific and modality-shared features using the modality discriminator and modality confuser, respectively. The modality discriminator effectively distinguishes between different modal features, endowing them with specific characteristics; whereas the modality confuser is unable to differentiate between modal features, thereby imparting shared characteristics to them. Since the modality-specific feature after previous stage contains substantial modality discrepancies, it is not suitable for direct distillation into the shared feature. 
We initially employ Instance Normalization to reduce domain discrepancies. However, it is crucial to acknowledge that IN inevitably results in the loss of some discriminative features. Therefore, we aim to reduce its modality style discrepancy while preserving identity-aware discriminative knowledge.
Subsequently, we distill this implicit knowledge into the modality-shared feature at both the feature-level through feature graph structure, and the semantic-level through the logit vector to enhance its distinctiveness.
Finally, an alignment loss is proposed to minimize modality discrepancy on enhanced modality-shared features.

The main contributions of this paper can be summarized as follows:
\begin{itemize}
	\item We propose the Implicit Discriminative Knowledge Learning (IDKL) network to utilize the discriminative knowledge implicit in the modality-specific features to enhance the upper bound of discriminative power for the modality-shared feature.
	\item To reduce the modality style discrepancy without losing discriminative information of modality-specific information, we propose an IN-guided Information Purifier (IP), which is supervised by the discrimination enhancing loss and discrepancy reducing loss.
	\item The novel TGSA loss is developed to distill the discriminative modality-specific information into modality-shared feature and mitigate inter-modality discrepancy of modality-shared feature sufficiently. The substantial experimental results demonstrate the superiority of our method.
\end{itemize}
\begin{figure*}
	\begin{center}
		\includegraphics[width=0.95\linewidth]{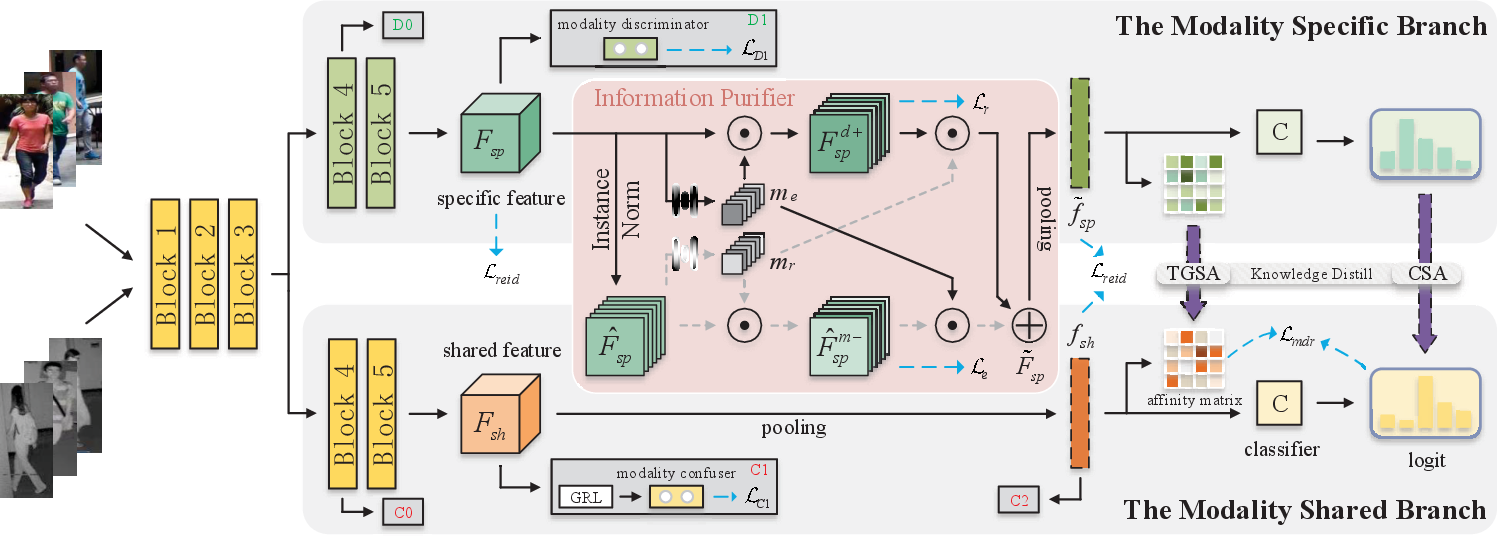}
	\end{center}
	\captionsetup{font=small}
	\caption{Framework of the proposed Implicit Discriminative Knowledge Learning (IDKL) model. The dual one-stream network built by resnet blocks first extracts the modality-specific $\boldsymbol{F}_{sp}$ and modality shared feature $\boldsymbol{F}_{sh}$ under the constraint of modality discriminator and modality confuser accordingly, while the common ReID loss is used to optimize network basely. Then, the modality-specific feature is fed into the information purifier to regulate the modality style discrepancy while preserving the implicit discriminative information and obtain the purified modality-specific feature $\widetilde{\boldsymbol{F}}_{sp}$. Subsequently, this implicit knowledge is distilled into the modality-shared feature through TGSA and CSA. Finally, the $\mathcal{L}_{mdr}$ is further proposed to minimize modality discrepancy within the enhanced modality-shared feature.}
	\label{fig:short}
\end{figure*}
\section{Related Work}
{\bf Visible Modality Person ReID.}
Person ReID has received increasing success in recent years, which aims to implement pedestrian retrieval between visible images. It has suffered huge challenges including various viewpoints, illuminations, postures and so on. To solve these problems, most methodologies \cite{9711131, 8382272, zhou2019omni,9336268} are designed to obtain unified intra-class and discriminative inter-class representation by training CNN network. To further enhance the distinguishability of features, Sun \etal~\cite{sun2018beyond} proposed aligning part feature directly instead of using external cues. 
He \etal~\cite{9710179} caught the hot spot of transformer and first proposed the transformer variant applying on single-modality person ReID.
Simultaneously, person ReID has developed some significant branches yet, such as Unsupervised Domain Adaptive Person ReID (UDA-ReID) and Domain Generalization Person ReID (DG-ReID) \cite{zhang2023style}. Although existing methods have made wide progress in visible modality person ReID, they suffer performance degradation when applied to visible-infrared person ReID due to the severe cross-modality discrepancy.

{\bf Visible-Infrared Person ReID (VI-ReID).}
VI-ReID is a cross-modality person retrieval problem, which aims at matching daytime visible and nighttime infrared images.
It not only faces difficulties encountered in traditional single-motality person ReID task, but also the main challenge of huge modality discrepancy which caused by different camera spectra. To solve the these issues, numerous approaches are proposed to search shared feature\cite{9745797, 9725265,  9336268, 9470916, 9879066, 8237837, ye2020dynamic,fu2022cross}.
Wu \etal~\cite{8237837} first formulated the VI-REID issue and contributed a new cross-modality dataset SYSU-MM01, which is of great significance to the following research.
Ye \etal~\cite{ye2020dynamic} proposed a part-level feature interaction and graph structure attention to enhance the discrimination of invariant feature.
Meanwhile, some work begin to introduce additional modal information to jointly search for invariant features \cite{9157347, 8953262, jiang2022cross, zhang2022modality, 9009814}.
Wang \etal~\cite{8953262} combined RGB three channels feature and IR single channel feature into united four channels feature to reduce modal differences.
Jiang \etal~\cite{jiang2022cross} employed a global prototype to generate the missing modality counterpart through transformer.
However, these generation based compensation methods inevitably introduce much noise and most of present methods aim to align modal discrepancy and extract modality invariant feature. They ignored use of beneficial style information contained in specific features to increase distinctiveness.

{\bf Interactive Learning.}
In knowledge distillation, a teacher-student model is used to transfer the knowledge learned by a larger complex teacher model to a smaller simple student model. Different from the one-way transfer between a teacher and a student, deep mutual learning \cite{zhang2018deep} is an ensemble of students
which learns collaboratively and teaches each other throughout the training process.
Some work applied this idea into VI-ReID for interactive learning in different modalities.
Zheng \etal~\cite{9935815} enhanced invariant features learning by using interactive learning between two modality.
Ye \etal~\cite{9107428} employed multi-classifier to reduce modality discrepancy on logits-level.
Wu \etal~\cite{9577837} distilled RGB and IR knowledge mutually by four classifier to achieve inter shared representation. 
However, they only learn the knowledge within the semantic-level.

\section{Methodology}
In this section, we detail the proposed Implicit Discriminative Knowledge Learning (IDKL) framework as shown in Fig.~\ref{fig:short}. IDKL first distinguishes modality-specific features from modality-shared features through two network branches, which are constrained by modality discriminator and modality confuser, respectively. Simultaneously, common ReID loss is used to endow feature with representation. Then, an Information Purifier (IP) is developed to reduce the impact of style variances while retaining identity-aware and discriminative knowledge in modality-specific features. Finally, we distill the implicit discriminative knowledge across two branches into the modality-shared feature through Triplet Graph Structure Alignment (TGSA) at the feature-level and Class Semantic Alignment (CSA) at the logit-level. Additionally, to reduce modality discrepancy of shared feature, we proposed the Modality Discrepancy Reduction (MDR) loss within the modality-shared branch. As the discrminative information increases and modality discrepancy decreases constantly, the enhanced modality-shared feature can be obtained.

Formally, we represent visible and infrared images in a dataset as $V=\{\boldsymbol{x}_i^V\}_{i=1}^{N_V}$ and $I=\{\boldsymbol{x}_i^I\}_{i=1}^{N_I}$, respectively. Typically, the number of images sampled in a mini-batch is equal across both modalities, i.e., $N_V = N_I = N = P \times K$, where $N_V$ and $N_I$ denote the number of images sampled from the visible and infrared modalities, respectively. Here, $N$ represents the number of images from a single modality, $2N$ is the total number of images in a mini-batch, $P$ is the number of distinct person classes, and $K$ is the number of images from each class within a single modality. We can therefore represent the images and their corresponding labels in a mini-batch as $X=\{\boldsymbol{x}_i | \boldsymbol{x}_i \in V \cup I\}_{i=1}^{2N}$ and $Y=\{\boldsymbol{y}_i\}_{i=1}^{N_p=P}$, respectively. These images $X$ are fed into two separate network branches to extract the modality-specific features $\boldsymbol{F}_{sp}$ which contains $\boldsymbol{F}_{sp,V},\boldsymbol{F}_{sp,I}$ and the shared features $\boldsymbol{F}_{sh}$ which contains $\boldsymbol{F}_{sh,V},\boldsymbol{F}_{sh,I}$ as follows:
\begin{equation}
	\boldsymbol{F}_{sp}=E_{sp}\left(\boldsymbol{x} \mid \Theta, \Psi\right), \boldsymbol{F}_{sh}=E_{sh}\left(\boldsymbol{x} \mid \Theta, \Phi \right),
\end{equation}
where $E_{sp}$ and $E_{sh}$ denote the specific and the shared feature extractor by employing ResNet-50, respectively. Among the sturctures of ResNet-50, the global average pooling replaced by Gem pooling\cite{8382272, 9336268} which is a pooling operation between maximum pooling and average pooling. And $\Theta$ is the shallow layer parameters with the first three blocks of ResNet-50, $\Psi$ and $\Phi$ is the deep layer parameters of different branches with the last two blocks of ResNet-50. 

\subsection{Modality Confuser and Discriminator}
\textbf{Modality Confuser.} For each sample image $\boldsymbol{x}_i$, there is a modality label $t_i \in \left\{0, 1\right\}$. To learn modality-irrelated shared information, similar to \cite{ganin2015unsupervised}, our goal is to confuse different domains such that a domain classifier cannot distinguish the domain of origin for a sample. We employ an adversarial modality classifier based on the Gradient Reversed Layer (GRL) as the modality confuser. The constraint loss for the modality confuser is given by:
\begin{equation}
	\mathcal{L}_{Cj}=-\frac{1}{2N} \sum_{i=1}^{2N} t_i \cdot \log p \left(C_{j}\left(GRL\left(\boldsymbol{F}_{sh}^i\right)\right)\right),
\end{equation}
where $C_{j}$ represents the j-th modality confuser in the modality-shared branch, $p\left(\cdot\right)$ is the prediction probability obtained via the softmax function, and $t_i$ is the modality label.

\textbf{Modality Discriminator.} To adequately learn modality-related specific information, we employ a modality classifier as the Modality Discriminator. This classifier, which does not use GRL, is applied on the specific branch to extract modality-specific features. The classification loss is formulated as follows:
\begin{equation}
	\mathcal{L}_{Dj}=-\frac{1}{2N} \sum_{i=1}^{2N} t_i \cdot \log p \left(D_{j}\left(\boldsymbol{F}_{sp}^i\right)\right),
\end{equation}
where $D_{j}$ denotes the j-th modality discriminator in the modality-specific branch.

The combined loss from the modality confuser and discriminator is given by:
\begin{equation}
	\mathcal{L}_{C}= \sum_{j=1}^{K} \mathcal{L}_{Cj}, \quad \mathcal{L}_{D}= \sum_{j=1}^{K} \mathcal{L}_{Dj}.
\end{equation}

To efficiently extract both modality-specific and modality-shared features, we combine these modality classifier losses with the standard ReID loss $\mathcal{L}_{reid}$, which includes cross-entropy and hard triplet loss. These are applied to both the modality-specific and modality-shared branches as follows:
\begin{equation}
	\begin{aligned}
		\mathcal{L}_{sp} = \mathcal{L}_{reid}\left(\boldsymbol{f}_{sp}\right) + \mathcal{L}_{D}, 
		\mathcal{L}_{sh} = \mathcal{L}_{reid}\left(\boldsymbol{f}_{sh}\right) + \mathcal{L}_{C},
	\end{aligned}
\end{equation}
where $\boldsymbol{f}\in\mathbb{R}^{B\times C }$ is the pooling feature corresponding to $\boldsymbol{F}\in\mathbb{R}^{B\times C \times H \times W}$.

Thus, the base loss for our model is formulated as:
\begin{equation}
	\begin{aligned}
		\mathcal{L}_{b} = \mathcal{L}_{sh} + \mathcal{L}_{sp}.
	\end{aligned}
\end{equation}

\subsection{Information Purifier}
The Information Purifier (IP) is designed to minimize the impact of style variances while retaining identity-aware and discriminative knowledge in modality-specific features. The IP integrates Instance Normalization (IN), which is known to reduce domain discrepancies \cite{jia2019frustratingly, zhou2019omni, pan2018two}. However, it is important to recognize that IN inevitably leads to the loss of some discriminative features \cite{huang2017arbitrary, jin2020style}, potentially hindering the high performance of ReID.

To overcome the aforementioned issues, we have designed a dual-mask network guided by Instance Normalization (IN) to alleviate modality style discrepancies while preserving implicit discriminative knowledge.
Firstly, we apply IN on the modality-specific feature to obtain the normalized feature $\widehat{\boldsymbol{F}}_{sp}$ by:
\begin{equation}
	\widehat{\boldsymbol{F}}_{sp}=\operatorname{IN}\left(\boldsymbol{F}_{sp}\right)=\frac{\boldsymbol{F}_{sp}-\mathrm{E}\left[\boldsymbol{F}_{sp}\right]}{\sqrt{\operatorname{Var}\left[\boldsymbol{F}_{sp}\right]+\epsilon}},
\end{equation}
where the $\epsilon$ represents a safety factor to ensure the denominator is not zero. The mean \( \text{E}[\cdot] \) and variance \( \text{Var}[\cdot] \) are calculated along each channel.

Following the approach of SE-Net \cite{hu2018squeeze}, we generate two channel-wise masks $\mathbf{m}_{e}$ and $\mathbf{m}_{r}$ by:
\begin{equation}
	\mathbf{m}_{e}=\sigma\left(\mathrm{W}_2 \delta\left(\mathrm{W}_1 g(\boldsymbol{F}_{sp})\right)\right),
	\mathbf{m}_{r}=\sigma(\mathrm{W}_4 \delta(\mathrm{W}_3 g(\widehat{\boldsymbol{F}}_{sp}))),
\end{equation}
where \( g(\cdot) \) denotes the pooling operation, \( W_1, W_3 \in \mathbb{R}^{\frac{c}{r} \times c} \)
and \( W_2, W_4 \in \mathbb{R}^{c \times \frac{c}{r}} \) are learnable parameters in the four fully-connected (FC) layers which are followed by the ReLU activation function \( \delta(\cdot) \) and the sigmoid activation function \( \sigma(\cdot) \). To balance the calculate consumption, the dimension reduction ratio \( r \) is set to 16.

The channel-wise masks $\mathbf{m}_{e}$ and $\mathbf{m}_{r}$ indicate enhancement of discriminative characteristics and reduction of discrepancies attention mask, respectively. So we can obtain the stronger distinctiveness $\boldsymbol{F}_{sp}^{d+}$ and the smaller modality differences $\widehat{\boldsymbol{F}}_{sp}^{m-}$ by:
\begin{equation}
	\boldsymbol{F}_{sp}^{d+}=\mathbf{m}_{e} \odot \boldsymbol{F}_{sp},\,\, \widehat{\boldsymbol{F}}_{sp}^{m-}=\mathbf{m}_{r} \odot \widehat{\boldsymbol{F}}_{sp}.
\end{equation}

Subsequently, the discrimination enhancing loss $\mathcal{L}_{e}$ and the discrepancy reducing loss $\mathcal{L}_{r}$ are calculated to supervise $\mathbf{m}_{e}$ and $\mathbf{m}_{r}$ respectively as:
\begin{equation}
	\mathcal{L}_{e}=\operatorname{Softplus}\left(h\left(C_{sp}\left(\boldsymbol{f}^{d+}_{sp}\right)\right)-h(C_{sp}(\boldsymbol{f}_{sp}))\right),
\end{equation}
\begin{equation}
	\mathcal{L}_{r}=\operatorname{Softplus}\left(d\left(\widehat{\boldsymbol{f}}_{sp,V}^{m-}, \widehat{\boldsymbol{f}}_{sp,I}^{m-}\right)-d\left(\widehat{\boldsymbol{f}}_{sp,V}, \widehat{\boldsymbol{f}}_{sp,I}\right)\right).
\end{equation}

Here, $\mathcal{L}_{e}$ aims to imbue the generated \(\mathbf{F}_{sp}^{d+}\) with greater semantic distinctiveness compared to \(\mathbf{F}_{sp}\), while $\mathcal{L}_{r}$ seeks to ensure that \(\widehat{\mathbf{F}}_{sp}^{m-}\) exhibit smaller modality differences than \(\widehat{\mathbf{F}}_{sp}\). And $\operatorname{Softplus}(\cdot)=\ln (1+\exp (\cdot))$ is a function with monotonic increase, designed to alleviate optimization challenges by circumventing negative values in loss.

Finally, an alternating integration strategy is employed to extract the distinctive information from \(\widehat{\mathbf{F}}_{sp}^{m-}\) by applying $\mathbf{m}_{e}$ and the invariant information from \(\mathbf{F}_{sp}^{d+}\) by applying $\mathbf{m}_{r}$. Thus, The two sets of features extracted in this way both  demonstrate smaller modality differences and stronger implicit discriminative information. By integrating them, we derive the purified modality-specific feature $\widetilde{\boldsymbol{F}}_{sp}$ as:
\begin{equation}
	\widetilde{\boldsymbol{F}}_{sp}=\mathbf{m}_{e} \odot \widehat{\boldsymbol{F}}_{sp}^{m-} + \mathbf{m}_{r} \odot \boldsymbol{F}_{sp}^{d+}.
\end{equation}

Intuitively, the pooling feature $\tilde{\boldsymbol{f}}_{sp}$ of implicit feature $\widetilde{\boldsymbol{F}}_{sp}$ is also constraint with ReID loss. So the information purify loss $\mathcal{L}_{ip}$ of this section is summerized as:
\begin{equation}
	\mathcal{L}_{ip}=\mathcal{L}_{e} + \mathcal{L}_{r} + \mathcal{L}_{reid}\left(\tilde{\boldsymbol{f}}_{sp}\right).
\end{equation}


\subsection{Implicit Knowledge Distillation}
To ensure that the modality-shared feature comprehensively learns and integrates implicit information, we perform distillation from both the feature-level through TGSA and logit-level through CSA.
\subsubsection{Triplet Graph Structure Alignment (TGSA)}
To endow the shared feature with discriminative information and reduce modality discrepancy at the feature level, we develop a triplet feature graph structure alignment loss. This approach is motivated by the fact that the feature graph structure contains abundant information about the relationships and distribution between features, such as inter-class distinctiveness and intra-class diversity. These characteristics are utilized to unearth potential feature relationships and enhance the feature representation in \cite{li2022counterfactual, ye2020dynamic}.
The graph structure affinity matrix which indicates the relationships between features is calculated by:
\begin{equation}
	\alpha_{ij}=\frac{\exp \left(L\left(\left[ l\left(\boldsymbol{f}_i\right) \Vert l\left(\boldsymbol{f}_j\right)\right] \cdot \boldsymbol{w}\right)\right)}{\sum_{k \in \mathcal{N}_i} \exp \left(L\left(\left[ l\left(\boldsymbol{f}_i\right) \Vert l\left(\boldsymbol{f}_k\right)\right] \cdot \boldsymbol{w}\right)\right)},
\end{equation}
\begin{figure}[t]
	\begin{center}
		\includegraphics[width=0.95\linewidth]{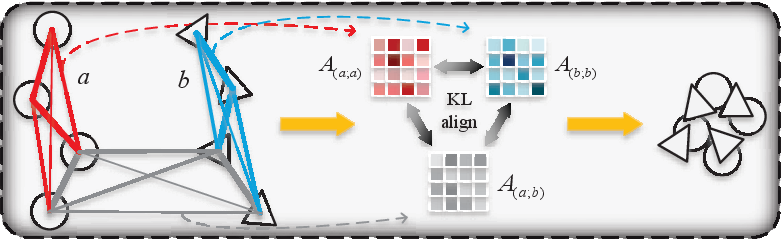}
	\end{center}
	\captionsetup{font=small}
	\caption{
		Illustration of the proposed TGSA loss: 'a' and 'b' denote two different types of features. '$\boldsymbol{A}$' represents the graph structure affinity matrix. 
		After aligning the three affinity matrices, the discrepancy in the graph structure distribution between features 'a' and features 'b' will be eliminated.
	}
	\label{fig:tgsa}
\end{figure}
where the $L$ denote LeakyReLU activation function, $\left[\cdot \Vert \cdot\right]$ denote concatenate operation and $\mathcal{N}_i$ denotes the neighbor samples which are used to normalize for the $i$th sample. $l\left(\cdot\right)$ is feature dimension transformation layer and $\boldsymbol{w}$ is the full connection layer to calculate scores in a pair features.
The affinity matrix is acquired making these scores through softmax function normalization. 

Due to we utilize graph structure to align and distill knowledge, instead of augmenting the feature, and the Euclidean space distribution is more interesting for features, thus we replace the linear transformation with Euclidean distance to compute attention scores 
and redefine the graph structure expression for two groups of features as follows:
\begin{equation}
		\boldsymbol{A}_{(a;b)}=\left\{\alpha_{ij}\right\}_{i,j\in N}
		=\frac{\exp \left(D\left(\boldsymbol{f}_{a}^i, \boldsymbol{f}_{b}^j\right)\right)}{\sum_{k \in N} \exp \left(D\left(\boldsymbol{f}_{a}^i, \boldsymbol{f}_{b}^k\right)\right)},
\end{equation}  
where $\alpha_{ij}$ denotes the element of the affinity matrix, $N$ corresponds to the entirety of samples within one modality, $a$ and $b$ represent two modalities, $D\left(\cdot\right)$ means Euclidean distance.

Specifically, the triplet graph structure alignment loss $\mathcal{L}_{tgsa}$ is developed for cross modality ReID to align two different modality types, enabling them to conform to the same graph structure distribution and reduce modality discrepancy. This loss encompasses two self-modal affinity matrices and one cross-modal affinity matrix, with the latter ensuring the overall consistency of the graph structure's distribution as shown in Fig.~\ref{fig:tgsa}.
These three matrices are aligned pairwise through the utilization of the Kullback-Leibler (KL) divergence. Therefore, the alignment loss $\mathcal{L}_{tgsa}$ of two distinct modality types is defined as:
\begin{equation}
\begin{aligned}
	\mathcal{L}_{tgsa}^{(a;b)}
	&=\sum_{p=1}^{P}\sum_{k=1}^{K} \left( \text{KL}\left(\boldsymbol{A}_{(a; a)}^{p_k}, \boldsymbol{A}_{(b; b)}^{p_k}\right)\right.\\
	&+\left.\text{KL}\left(\boldsymbol{A}_{(a; a)}^{p_k}, \boldsymbol{A}_{(a; b)}^{p_k}\right) + \text{KL}\left(\boldsymbol{A}_{(a;b)}^{p_k}, \boldsymbol{A}_{(b; b)}^{p_k}\right) \right).
\end{aligned}
\end{equation}

Here, $\boldsymbol{A}^{p_k}$ represents the graph structure distribution of the $k$-th sample in the $p$-th class, where $P$ is the number of person classes, and $K$ denotes the number of images for each class within a single modality.

In order to convey the discriminative implicit modality-specific knowledge to shared feature on feature-level, the distillation loss across two branches on homogeneous features through TGSA can be formulated as:
\begin{equation}
	\mathcal{L}_{tgsa}=\mathcal{L}_{tgsa}^{(sp,V; sh,V)} + \mathcal{L}_{tgsa}^{(sp,I; sh,I)}.
\end{equation}


\subsubsection{Class Sementic Alignment (CSA) }
CSA is used to distill the semantic information of implicit modality-specific knowledge into the modality-shared branch for enhancing the feature representation of shared features. CSA operates on homogeneous features between two branches at logit-level. 
The logit matrix behind the classifier 
can be formulated as:
\begin{equation}	
\boldsymbol{Z}_{sp}= C_{sp}\left(\boldsymbol{f}_{sp}\right),
\boldsymbol{Z}_{sh}= C_{sh}\left(\boldsymbol{f}_{sh}\right),
\end{equation}
where $C_{sp}$ and $C_{sh}$ is the modality-specific and modality-shared classifier separately. And the logit $\boldsymbol{Z}_{sp}$ and $\boldsymbol{Z}_{sh} \in \mathbb{R}^{2N\times C}$ of specific and shared branch both contain visible modality and infrared modality, $C$ is the total number of train dataset identities.

For learning the implicit discriminative modality-specific knowledge on semantic-level, the CSA loss is implemented on the same modality logit between the two branches, which is formulated as:
\begin{equation}
	\mathcal{L}_{csa}=\sum_{i=1}^N  \left(\text{KL} \left(\boldsymbol{Z}_{sh,V}^{i}, \boldsymbol{Z}_{sp,V}^{i}\right) + \text{KL} \left(\boldsymbol{Z}_{sh,I}^{i}, \boldsymbol{Z}_{sp,I}^{i}\right)\right).
\end{equation}

\subsection{Modality Discrepancy Reduction (MDR) }
At this part, to guarantee the invariant representation of modality-shared feature, the TGSA and CSA are further used to reduce modality discrepancy within modality-shared branch as follow:
\begin{equation}
\mathcal{L}_{mdr}= \mathcal{L}_{tgsa}^{(sh,V; sh,I)}+\sum_{i=1}^N \text{KL} \left(\boldsymbol{Z}_{sh,V}^{i}, \boldsymbol{Z}_{sh,I}^{i}\right).
\end{equation}

In this way, the visible feature and infrared feature in modality-shared branch can achieve mutual learning from feature-level and semantic-level. It makes two modalities feature aligning information each other, while alleviating modal gap and maintaining the invariant of modality-shared feature.
\subsection{Optimization}
Ultimately, by continuously distilling implicit discriminative knowledge from the modality-specific feature and consistently reducing modality discrepancies in the modality-shared feature, we can achieve a more discriminative and invariant modality-shared feature.

The total loss of the model IDKL is defined as:
\begin{multline}
\mathcal{L}_{total}= \mathcal{L}_{b}+\lambda_1\mathcal{L}_{ip}+\lambda_2 \mathcal{L}_{tgsa} + \lambda_3 \mathcal{L}_{csa} + \mathcal{L}_{mdr}, \\
\end{multline}
where $\lambda_1$, $\lambda_2$ and $\lambda_3$ are hype-parameters to balance the
contribution of individual loss term.

\section{Experiments}
\subsection{Datasets and Experimental Settings}
\textbf{Datasets.}
Three public VI-ReID datasets SYSU-MM01 \cite{9009814}, LLCM \cite{zhang2023diverse} and RegDB \cite{nguyen2017person} are employed to evaluate our model. SYSU-MM01 is a popular large-scale dataset collected by four
visible cameras and two near-infrared cameras, including indoor and outdoor environments. And the test protocols consist of all-search and indoor-search.
LLCM dataset is a large-scale and low-light cross-modality dataset, which is divided into training and testing sets at a 2:1 ratio. 
RegDB is collected using dual-camera systems, where visible and infrared images are captured in pairs.
Both LLCM and RegDB contain visible to infraed and infrared to visible two search modes.

\textbf{Evaluation metrics.} 
The standard rank-$k$ matching accuracy and mean Average Precision
(mAP) are adopted as the evaluation metrics. All the reported results are the average of 10 trials.

\textbf{Implementation details.} The proposed method and all experiments are implemented on a single NVIDIA GeForce 3090 GPU with PyTorch framework. The baseline model adopts the ResNet-50 pre-trained on ImageNet with the $\mathcal{L}_{b}$.  The input images are resized to $3\times384\times128$. The train mini-batch size is set as 120, which contains 12 random identities and 10 images for every identity. Adam optimizer with an initial learning rate $3\times10^{-5}$ is exploited, which decays at the $60$th and $100$th epoch with a decay factor of $0.1$. The hype-parameters $\lambda_1$, $\lambda_2$ and $\lambda_3$ are set to $0.1$, $0.6$ and $0.8$.
During the testing phase, only modality-shared feature is used to evaluate performance.
\begin{figure}[t]
\begin{center}
\includegraphics[width=0.95\linewidth]{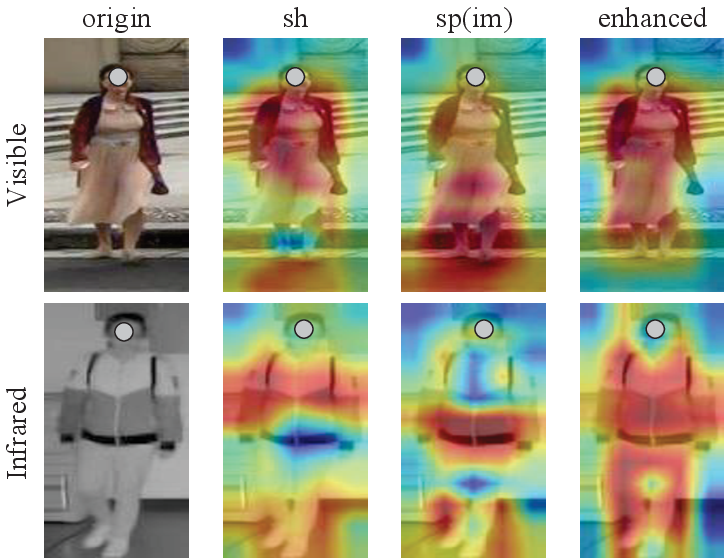}
\end{center}
\captionsetup{font=small}
\vspace{-5pt} 
\caption{ Observation the implicit discriminative information by Grad-CAM. And 'sh' and 'sp(im)' present the modality-shared feature and the modality-specific feature of trained IDKL w/o knowledge distillation, respectively; 'enhanced' denotes the modality-shared feature of IDKL w/ knowledge distillation.
}
\label{fig:cam}
\vspace{5pt} 
\end{figure}
\begin{table*}[t]
\begin{center}
\caption{\small Comparison of CMC (\%) and mAP (\%) performances with the state-of-the-art methods on \textbf{SYSU-MM01} dataset.}
\resizebox{\linewidth}{!}{
	\begin{tabular}{l||ccc|c|ccc|c||ccc|c|ccc|c}
		\hline \multirow{3}{*}{Methods}&\multicolumn{8}{c||}{All-search}&\multicolumn{8}{c}{Indoor-search}\\
		\cline {2 - 17}&\multicolumn{4}{c|}{Single-shot}&\multicolumn{4}{c||}{Multi-shot}&\multicolumn{4}{c|}{Single-shot}&\multicolumn{4}{c}{Multi-shot}\\
		\cline {2 - 17}&r=1&r=10&r=20&map&r=1&r=10&r=20&map&r=1&r=10&r=20&map&r=1&r=10&r=20&map\\
		\cline {1 - 17}
		Zero-Padding \cite{8237837}&14.80&54.12 &71.33 &15.95 &19.13& 61.40 &78.41 &10.89 &20.58 &68.38& 85.79 &26.92& 24.43&75.86 &91.32& 18.86\\
		D-HSME \cite{hao2019hsme}&20.68 &62.74 &77.95 &23.12&- &- &- &- &-& - &-& - &-& -& -& -\\
		AlignGAN \cite{9009814}& 42.40 &85.00 &93.70& 40.70& 51.50& 89.40& 95.70 &33.90 &45.90 &87.60& 94.40& 54.30 &57.10 &92.70& 97.40 &45.30\\
		DDAG \cite{ye2020dynamic}&54.75 &90.39 &95.81 &53.02&-&-&-&-&61.02 &94.06 &98.41& 67.98&-&-&-&-\\
		
		NFS \cite{9577446}& 56.91 &91.34 &96.52 &55.45 &63.51& 94.42 &97.81 &48.56& 62.79& 96.53& 99.07& 69.79 &70.03 &97.70 &99.51 &61.45\\
		PIC \cite{9935815}&57.51& 89.35 &95.03& 55.14&- &- &- &- &60.40& - &-&67.70 &-& -& -& -\\
		MID \cite{huang2022modality}&60.27&92.90&-&59.40 &-&-&-&-&64.86 &96.12 &-&70.12&-&-&-&-\\
		cm-SSFT \cite{9157347} &61.60&89.20&93.90&63.20&63.40&91.20&95.70&62.00&70.50&94.90&97.70& 72.60&73.00&96.30&99.10&72.40\\
		MCLNet \cite{9711447} &65.40 &93.33& 97.14& 61.98&-&-&-&-&72.56 &96.98 &99.20 &76.58&-&-&-&-\\
		FMCNet \cite{9880449} &66.34&-&-&62.51&73.44&-&-&56.06&68.15&-&-&74.09&78.86&-&-&63.82\\
		SMCL \cite{9711501} &67.39&92.87&96.76&61.78&72.15&90.66&94.32&54.93&68.84&96.55&98.77&75.56& 79.57&95.33&98.00&66.57\\
		CAJ \cite{9711494}&69.88&95.71&98.4&66.89&-&-&-&-&76.26&97.88&99.49&80.37&-&-&-&-\\
		MPANet \cite{9577837}& 70.58& 96.21& 98.80& 68.24& 75.58& 97.91 &99.43 &62.91 &76.74& 98.21& 99.57& 80.95& 84.22 &99.66& 99.96& 75.11\\
		CMT \cite{jiang2022cross}&71.88&96.45& 98.87 &68.57 &80.23& 97.91&99.53& 63.13& 76.9 &97.68 &99.64 &79.91& 84.87 &99.41& $\textbf{99.97}$ &74.11\\
		DEEN \cite{zhang2023diverse} &74.7& 97.6& \textbf{99.2}& 71.8&-&-&-&-& 80.3& 99.0& 99.8 &83.3&-&-&-&-\\
		SAAI \cite{fang2023visible} &75.90 &-&-&77.03 &82.8&-&-& $\textbf{82.39}$& 83.20 &-&-&88.01& 90.73&-&-& $\textbf{91.30}$\\
		MUN \cite{yu2023modality}&76.24 &\textbf{97.84} &-&73.81& -&-&-&-&79.42 &98.09&-& 82.06&-&-&-&-\\
		MSCLNet \cite{zhang2022modality} &76.99 &97.63 &99.18& 71.64&-&-&-&-&78.49 &\textbf{99.32} &\textbf{99.91} &81.17& -&-&-&-\\
		PartMix \cite{kim2023partmix}&77.78&-&-& 74.62& 80.54&-&-& 69.84& 81.52&-&-& 84.38& 87.99&-&-& 79.95\\
		\cline {1 - 17}
		IDKL(Ours)&$\textbf{81.42}$ &97.38&98.89&$\textbf{79.85}$&\textbf{84.34} &\textbf{98.89} &\textbf{99.73}&78.22&$\textbf{87.14}$&98.28&99.26&$\textbf{89.37}$&$\textbf{94.30}$&$\textbf{99.71}$&99.93&88.75\\
		\hline
	\end{tabular}
}
\label{SYSU}
\end{center}
\vspace{-11pt}
\end{table*}

\subsection{Comparison with State-of-the-art Methods}
We compare our IDKL model with the state-of-the-art VI-ReID
methods published recent years on public VI-ReID datasets SYSU-MM01, RegDB and LLCM.

\textbf{Comparison on SYSU-MM01 dataset.} 
The comparison experimental results is shown in \cref{SYSU} which displays the proposed IDKL method outperforms existing cutting-edge methods. Specifically, the IDKL method achieves the accuracy of 81.42\% rank-1 and 79.85\% map with single-shot all search protocol
, while the accuracy of 87.14\% rank-1 and 89.37\% map with single-shot indoor search protocol.
The compared SOTAs include various base methods, \ie, for the methods of learning the shared feature through network and loss function immediately, which contain D-HSME 
\cite{hao2019hsme}, NFS \cite{9577446}
, CMT \cite{jiang2022cross}, DEEN\cite{zhang2023diverse} and MCLNet \cite{9711447}.
Based on graph structure augment method DDAG \cite{ye2020dynamic} and the mutual learning by logits skill (PIC \cite{9935815} and MPANet \cite{9577837}). Comparing with several other based modality-specific methods cm-SSFT\cite{9157347} , MUN\cite{yu2023modality} and MSCLNet\cite{zhang2022modality}, our results outperform them by a margin. 

\begin{figure}[t]
\begin{center}
\includegraphics[width=1\linewidth]{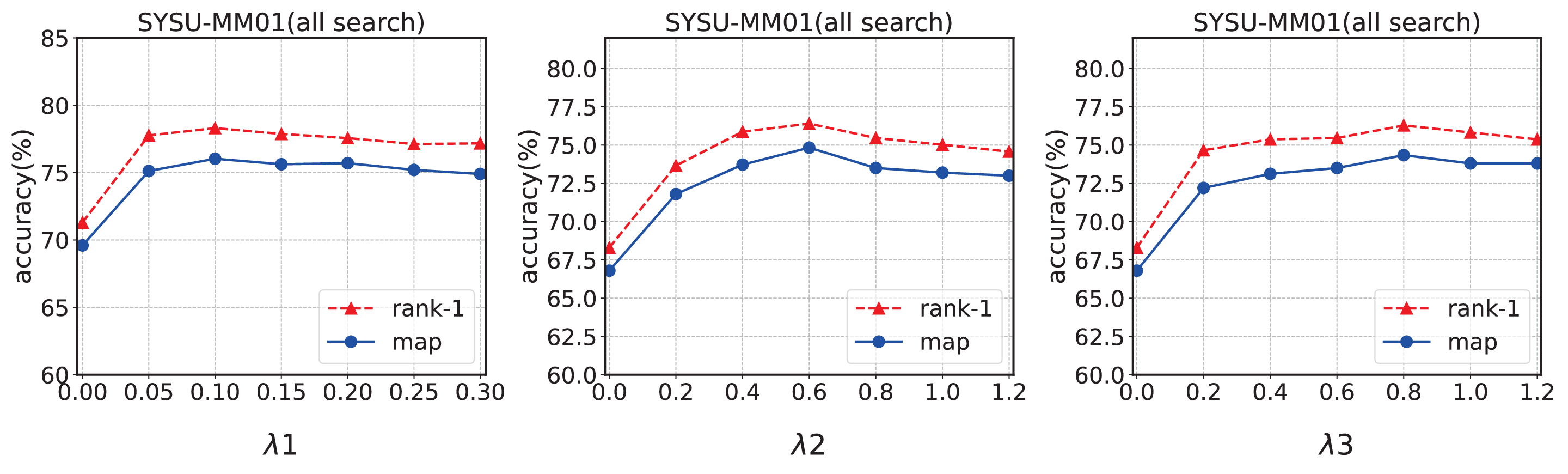}
\end{center}
\captionsetup{font=small}
\vspace{-9pt} 
\caption{Ablation analysis of hyper-parameter $\lambda_1$ and $\lambda_2$, $\lambda_3$ for $\mathcal{L}_{ip}$, $\mathcal{L}_{tgsa}$, and $\mathcal{L}_{csa}$ respectively on SYSU-MM01 dataset.
}
\label{fig:line}
\vspace{0pt}
\setlength{\textfloatsep}{2pt}
\end{figure}
\textbf{Comparison on RegDB dataset.} 
We also evaluate IDKL on
a small-scale dataset RegDB as shown in  \cref{RegDB}. There is strong  performance IDKL showed and outperforms the existing solutions. 
Specifically, we achieve rank-1 accuracy of 94.72\% in visible to infrared mode, and rank-1
accuracy of 94.22\%  in infrared to visible mode.

\textbf{Comparison on LLCM dataset.} The IDKL model achieves significant improvements on the large and complex LLCM dataset as shown in \cref{LLCM}, demonstrating excellent rank-1 accuracy of 72.2\% and 70.7\% on two modes, respectively. This indicates that the IDKL model exhibits strong robustness in complex and multimodal scenarios.

\begin{table}[htbp]
\begin{center}
\vspace{0pt}
\caption{\small Comparison of the CMC (\%) and mAP (\%) performances
	with state-of-the-art methods on \textbf{RegDB} dataset.}
\resizebox{\linewidth}{!}{
	\begin{tabular}{l||cc|cc}
		\hline \multirow{2}{*}{ Methods } & \multicolumn{2}{c|}{ Visible to infrared } & \multicolumn{2}{c}{ Infrared to visible } \\
		\cline { 2 - 5 }&rank-1&map&rank-1&map \\
		\hline 
		Zero-Padding \cite{8237837}& 17.8& 18.9& 16.7 &17.9\\
		AlignGAN \cite{9009814} &57.9 &53.6& 56.3 &53.4\\
		DDAG \cite{ye2020dynamic}& 69.34&63.46&68.06&61.80\\
		cm-SSFT \cite{9157347}&72.3& 72.9& 71.0 &71.7 \\
		MCLNet \cite{9711447}&80.31&73.07&75.93&69.49\\
		PIC \cite{9935815}&83.6 &79.6&79.5 &77.4\\
		MPANet \cite{9577837} &83.7 &80.9& 82.8 &80.7\\
		SMCL \cite{9711501}&83.93 &79.83&83.05& 78.57\\
		MSCLNet \cite{zhang2022modality} &84.17 &80.99 &83.86 &78.31\\
		CAJ \cite{9711494}&85.03&77.82&84.75&77.82\\
		MID \cite{huang2022modality}&87.45&84.85&84.29&81.41\\
		FMCNet \cite{9880449}&89.12& 84.43&88.38 &83.86\\
		SAAI \cite{fang2023visible} & 91.07& 91.45&92.09& 92.01\\
		DEEN \cite{zhang2023diverse}&91.1&85.1 &89.5&83.4\\
		
		CMT \cite{jiang2022cross} &95.17&87.3&91.97&84.46\\
		MUN \cite{yu2023modality}  &\textbf{95.19} & 87.15&91.86 &85.01\\

		\hline IDKL(Ours)&94.72&$\textbf{90.19}$&$\textbf{94.22}$&$\textbf{90.43}$\\
		\hline
	\end{tabular}
}
\label{RegDB}
\end{center}
\vspace{-18pt}
\end{table}
\begin{table}[htbp]
\vspace{-10pt}
\begin{center}
\caption{\small Comparison of the CMC (\%) and mAP (\%) performances
	with state-of-the-art methods on \textbf{LLCM} dataset.}
\resizebox{\linewidth}{!}{
	\begin{tabular}{l||cc|cc}
		\hline \multirow{2}{*}{ Methods } & \multicolumn{2}{c|}{ Visible to infrared } & \multicolumn{2}{c}{ Infrared to visible } \\
		\cline { 2 - 5 }&rank-1&map&rank-1&map \\
		\hline 
		DDAG \cite{ye2020dynamic}&40.3 & 48.4&48.0& 52.3\\
		CAJ \cite{9711494} &56.5& 59.8&48.8& 56.6\\
		DEEN \cite{zhang2023diverse} &62.5 &65.8&54.9 &62.9\\
		
		\hline IDKL(Ours)&$\textbf{72.22}$&$\textbf{66.43}$&$\textbf{70.72}$&$\textbf{65.19}$\\
		\hline
	\end{tabular}
}
\label{LLCM}
\end{center}
\vspace{-5pt}
\end{table}
\begin{table}[htbp]
\vspace{-8pt}
\begin{center}
\caption{\small Evaluation the impact of different components in terms of rank-1 (\%) and mAP (\%) on \textbf{SYSU-MM01} dataset.}
\begin{tabular}{ccccc||cc}
	\toprule
	\multirow{2}{*}{$\mathcal{L}_{b}$}&\multirow{2}{*}{$\mathcal{L}_{ip}$}&\multirow{2}{*}{$\mathcal{L}_{tgsa}$}&\multirow{2}{*}{$\mathcal{L}_{csa}$}&\multirow{2}{*}{$\mathcal{L}_{mdr}$} & \multicolumn{2}{c}{SYSU-MM01} \\
	\cmidrule(r){6-7} & & & & &  \multicolumn{1}{c}{rank1} & map \\
	\midrule
	$\checkmark$&$\times$&$\times$& $\times$& $\times$& 67.60 & 66.47 \\
	$\checkmark$&$\times$&$\checkmark$& $\times$& $\times$&68.38 & 67.30  \\
	$\checkmark$&$\checkmark$&$\checkmark$&$\times$&$\times$& 76.40 & 74.83\\
	$\checkmark$&$\checkmark$&$\times$&$\checkmark$& $\times$& 76.28 & 74.34 \\
	$\checkmark$&$\checkmark$&$\times$&$\times$&$\checkmark$& 77.04 & 75.66 \\
	$\checkmark$&$\checkmark$&$\checkmark$&$\checkmark$&$\checkmark$& 81.42 & 79.85 \\
	\bottomrule
\end{tabular}
\label{ablation}
\end{center}
\vspace{-5pt} 
\end{table}

\subsection{Ablation Study}
In this subsection, we conduct the ablation experiment to
evaluate our proposed model exhaustively.

\textbf{Effectiveness of each component.}
We evaluate the effectiveness of each component on the SYSU-MM01 dataset under all search  single-shot mode. Each component is added independently to reveal its the performance as \cref{ablation}.
This indicates that each component is highly useful, with $\mathcal{L}_{mdr}$
achieving significant performance improvements. This further suggests the effectiveness of TGSA and CSA in reducing modality discrepancies. Comparing the second row with the third row demonstrates the necessity of purifying implicit modality-specific information, and also proves the effectiveness of our Information Purifier module.  

\textbf{Hyper-parameters analysis of IP, TGSA and CSA.}
In this part, we present a line chart to examine the detail influence of IP, TGSA and CSA by gradually increasing the value of hyperparameters.
As depicted in Fig.~\ref{fig:line}, the maximum contributions of IP, TGSA and CSA are reached at 0.1, 0.6 and 0.8, respectively. The upward trend of the curve demonstrates the effectiveness of each module. 

\begin{figure}[t]
\begin{center}
\includegraphics[width=1\linewidth]{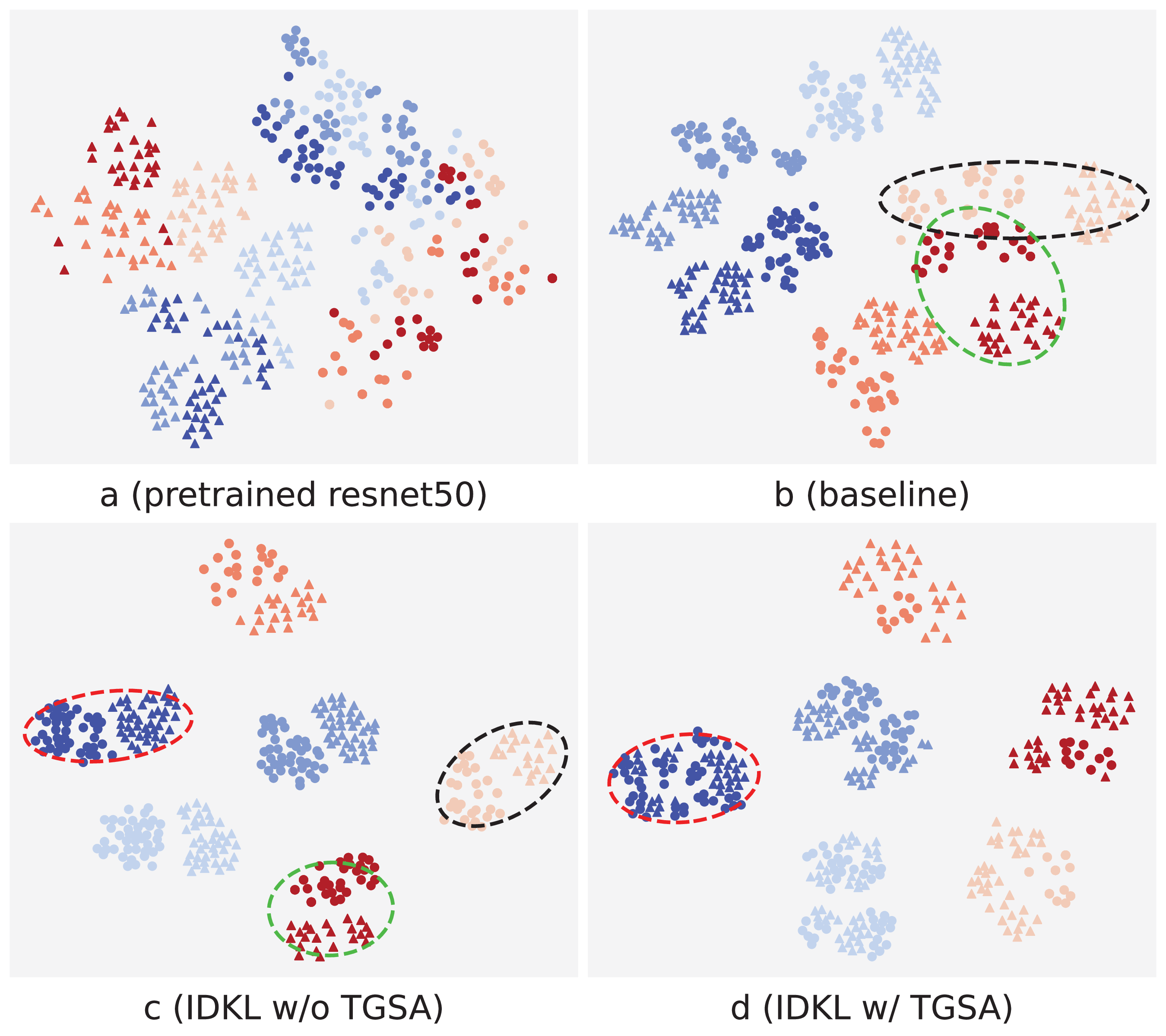}
\end{center}
\captionsetup{font=small}
\vspace{-2pt}
\caption{Visualization of learned features by t-SNE. 
Where "a" means the pre-training resnet50 on ImageNet; "b" represents the baseline; "c" is the IDKL model w/o TGSA; "d" is IDKL model w/ TGSA.}

\label{fig:tsne}
\end{figure}

\subsection{Visualization Analysis}

\textbf{Attention maps Visualization.}
To further illustrate the effectiveness of IDKL, Grad-CAM \cite{selvaraju2017grad} is utilized for a visual examination of different feature heatmaps. In Fig.~\ref{fig:cam}, the areas of focus for implicit discriminative information typically differ from those of the shared feature, indicating that effectively and judiciously utilizing this information to strengthen the shared feature can be highly beneficial.

\textbf{Feature Distribution Visualization.}
We utilize t-SNE \cite{nguyen2017person} feature map visualization to observe the impact of the IDKL model and TGSA on the model. As shown in Fig.~\ref{fig:tsne}, each color represents a different identity, while the shapes of circles and triangles indicate the visible and infrared modality information, respectively. 
From Fig.~\ref{fig:tsne}(a), a significant discrepancy between the two modalities is evident. Fig.~\ref{fig:tsne}(b) shows a reduction in modality discrepancy and the model exhibiting some discriminative capabilities. Comparing Fig.~\ref{fig:tsne}(b) with Fig.~\ref{fig:tsne}(c), it is observable that the IDKL model has a smaller intra-class discrepancy and better inter-class discrimination. In Fig.~\ref{fig:tsne}(d), the dark blue class appears more scattered than in Fig.~\ref{fig:tsne}(c), and the graph structures of the two modalities are more similar and closely aligned, demonstrating the effectiveness of TGSA in reducing modality discrepancies.

\section{Conclusion}

This paper harnesses the implicit discriminative information within modality-specific features and introduces a robust model IDKL to exploit the potential discrimination of heterogeneous-related features and enhance the shared feature. The IDKL model comprises a dual one-stream network, a novel IN-guided information purifier, a triplet graph structure alignment solution, and refined distillation on logits. Collectively, these components demonstrate exceptional effectiveness and contribute to improved results.
\section*{Acknowledgments}
This work was partially supported by National Key R\&D Program of China (2021YFB3100800), National Natural Science Fund of China (62271090, 61771079), Chongqing Natural Science Fund (cstc2021jcyj-jqX0023) and National Youth Talent Project. This work is also supported by Huawei computational power of Chongqing Artificial Intelligence Innovation Center.

{
	\small
	\bibliographystyle{ieeenat_fullname}
	\bibliography{main}

\begin{thebibliography}{49}
\providecommand{\natexlab}[1]{#1}
\providecommand{\url}[1]{\texttt{#1}}
\expandafter\ifx\csname urlstyle\endcsname\relax
  \providecommand{\doi}[1]{doi: #1}\else
  \providecommand{\doi}{doi: \begingroup \urlstyle{rm}\Url}\fi

\bibitem[Chen et~al.(2022)Chen, Ye, Qi, Wu, Jiang, and Lin]{9725265}
Cuiqun Chen, Mang Ye, Meibin Qi, Jingjing Wu, Jianguo Jiang, and Chia-Wen Lin.
\newblock Structure-aware positional transformer for visible-infrared person
  re-identification.
\newblock \emph{IEEE Transactions on Image Processing}, 31:\penalty0
  2352--2364, 2022.

\bibitem[Chen et~al.(2021)Chen, Wan, Li, Jing, and Sun]{9577446}
Yehansen Chen, Lin Wan, Zhihang Li, Qianyan Jing, and Zongyuan Sun.
\newblock Neural feature search for rgb-infrared person re-identification.
\newblock In \emph{2021 IEEE/CVF Conference on Computer Vision and Pattern
  Recognition (CVPR)}, pages 587--597, 2021.

\bibitem[Fang et~al.(2023)Fang, Yang, and Fu]{fang2023visible}
Xingye Fang, Yang Yang, and Ying Fu.
\newblock Visible-infrared person re-identification via semantic alignment and
  affinity inference.
\newblock In \emph{Proceedings of the IEEE/CVF International Conference on
  Computer Vision}, pages 11270--11279, 2023.

\bibitem[Fu et~al.(2021)Fu, Hu, Wu, Shi, Mei, and He]{9710774}
Chaoyou Fu, Yibo Hu, Xiang Wu, Hailin Shi, Tao Mei, and Ran He.
\newblock Cm-nas: Cross-modality neural architecture search for
  visible-infrared person re-identification.
\newblock In \emph{2021 IEEE/CVF International Conference on Computer Vision
  (ICCV)}, pages 11803--11812, 2021.

\bibitem[Fu et~al.(2022)Fu, Huang, Zhou, Ma, Xu, and Zhang]{fu2022cross}
Xiaowei Fu, Fuxiang Huang, Yuhang Zhou, Huimin Ma, Xin Xu, and Lei Zhang.
\newblock Cross-modal cross-domain dual alignment network for rgb-infrared
  person re-identification.
\newblock \emph{IEEE Transactions on Circuits and Systems for Video
  Technology}, 32\penalty0 (10):\penalty0 6874--6887, 2022.

\bibitem[Ganin and Lempitsky(2015)]{ganin2015unsupervised}
Yaroslav Ganin and Victor Lempitsky.
\newblock Unsupervised domain adaptation by backpropagation.
\newblock In \emph{International conference on machine learning}, pages
  1180--1189. PMLR, 2015.

\bibitem[Hao et~al.(2021)Hao, Zhao, Ye, and Shen]{9711447}
Xin Hao, Sanyuan Zhao, Mang Ye, and Jianbing Shen.
\newblock Cross-modality person re-identification via modality confusion and
  center aggregation.
\newblock In \emph{2021 IEEE/CVF International Conference on Computer Vision
  (ICCV)}, pages 16383--16392, 2021.

\bibitem[Hao et~al.(2019)Hao, Wang, Li, and Gao]{hao2019hsme}
Yi Hao, Nannan Wang, Jie Li, and Xinbo Gao.
\newblock Hsme: Hypersphere manifold embedding for visible thermal person
  re-identification.
\newblock In \emph{Proceedings of the AAAI conference on artificial
  intelligence}, pages 8385--8392, 2019.

\bibitem[He et~al.(2021)He, Luo, Wang, Wang, Li, and Jiang]{9710179}
Shuting He, Hao Luo, Pichao Wang, Fan Wang, Hao Li, and Wei Jiang.
\newblock Transreid: Transformer-based object re-identification.
\newblock In \emph{2021 IEEE/CVF International Conference on Computer Vision
  (ICCV)}, pages 14993--15002, 2021.

\bibitem[Hu et~al.(2018)Hu, Shen, and Sun]{hu2018squeeze}
Jie Hu, Li Shen, and Gang Sun.
\newblock Squeeze-and-excitation networks.
\newblock In \emph{Proceedings of the IEEE conference on computer vision and
  pattern recognition}, pages 7132--7141, 2018.

\bibitem[Huang and Belongie(2017)]{huang2017arbitrary}
Xun Huang and Serge Belongie.
\newblock Arbitrary style transfer in real-time with adaptive instance
  normalization.
\newblock In \emph{Proceedings of the IEEE international conference on computer
  vision}, pages 1501--1510, 2017.

\bibitem[Huang et~al.(2022)Huang, Liu, Li, Zheng, and Zha]{huang2022modality}
Zhipeng Huang, Jiawei Liu, Liang Li, Kecheng Zheng, and Zheng-Jun Zha.
\newblock Modality-adaptive mixup and invariant decomposition for rgb-infrared
  person re-identification.
\newblock In \emph{Proceedings of the AAAI Conference on Artificial
  Intelligence}, pages 1034--1042, 2022.

\bibitem[Jia et~al.(2019)Jia, Ruan, and Hospedales]{jia2019frustratingly}
Jieru Jia, Qiuqi Ruan, and Timothy~M Hospedales.
\newblock Frustratingly easy person re-identification: Generalizing person
  re-id in practice.
\newblock \emph{arXiv preprint arXiv:1905.03422}, 2019.

\bibitem[Jiang et~al.(2022)Jiang, Zhang, Liu, Qian, Zhang, and
  Wu]{jiang2022cross}
Kongzhu Jiang, Tianzhu Zhang, Xiang Liu, Bingqiao Qian, Yongdong Zhang, and
  Feng Wu.
\newblock Cross-modality transformer for visible-infrared person
  re-identification.
\newblock In \emph{Computer Vision--ECCV 2022: 17th European Conference, Tel
  Aviv, Israel, October 23--27, 2022, Proceedings, Part XIV}, pages 480--496.
  Springer, 2022.

\bibitem[Jin et~al.(2020)Jin, Lan, Zeng, Chen, and Zhang]{jin2020style}
Xin Jin, Cuiling Lan, Wenjun Zeng, Zhibo Chen, and Li Zhang.
\newblock Style normalization and restitution for generalizable person
  re-identification.
\newblock In \emph{proceedings of the IEEE/CVF conference on computer vision
  and pattern recognition}, pages 3143--3152, 2020.

\bibitem[Kim et~al.(2023)Kim, Kim, Park, Park, and Sohn]{kim2023partmix}
Minsu Kim, Seungryong Kim, Jungin Park, Seongheon Park, and Kwanghoon Sohn.
\newblock Partmix: Regularization strategy to learn part discovery for
  visible-infrared person re-identification.
\newblock In \emph{Proceedings of the IEEE/CVF Conference on Computer Vision
  and Pattern Recognition}, pages 18621--18632, 2023.

\bibitem[Li et~al.(2022)Li, Lu, Liu, Liu, Yin, Chu, Huang, Zhu, Zhao, and
  Yu]{li2022counterfactual}
Xulin Li, Yan Lu, Bin Liu, Yating Liu, Guojun Yin, Qi Chu, Jinyang Huang, Feng
  Zhu, Rui Zhao, and Nenghai Yu.
\newblock Counterfactual intervention feature transfer for visible-infrared
  person re-identification.
\newblock In \emph{Computer Vision--ECCV 2022: 17th European Conference, Tel
  Aviv, Israel, October 23--27, 2022, Proceedings, Part XXVI}, pages 381--398.
  Springer, 2022.

\bibitem[Liang et~al.(2021)Liang, Wang, Lai, and Xie]{9470916}
Wenqi Liang, Guangcong Wang, Jianhuang Lai, and Xiaohua Xie.
\newblock Homogeneous-to-heterogeneous: Unsupervised learning for rgb-infrared
  person re-identification.
\newblock \emph{IEEE Transactions on Image Processing}, 30:\penalty0
  6392--6407, 2021.

\bibitem[Liu et~al.(2021)Liu, Tan, and Zhou]{9276429}
Haijun Liu, Xiaoheng Tan, and Xichuan Zhou.
\newblock Parameter sharing exploration and hetero-center triplet loss for
  visible-thermal person re-identification.
\newblock \emph{IEEE Transactions on Multimedia}, 23:\penalty0 4414--4425,
  2021.

\bibitem[Liu et~al.(2022)Liu, Sun, Zhu, Pei, Yang, and Li]{9879066}
Jialun Liu, Yifan Sun, Feng Zhu, Hongbin Pei, Yi Yang, and Wenhui Li.
\newblock Learning memory-augmented unidirectional metrics for cross-modality
  person re-identification.
\newblock In \emph{2022 IEEE/CVF Conference on Computer Vision and Pattern
  Recognition (CVPR)}, pages 19344--19353, 2022.

\bibitem[Lu et~al.(2020)Lu, Wu, Liu, Zhang, Li, Chu, and Yu]{9157347}
Yan Lu, Yue Wu, Bin Liu, Tianzhu Zhang, Baopu Li, Qi Chu, and Nenghai Yu.
\newblock Cross-modality person re-identification with shared-specific feature
  transfer.
\newblock In \emph{2020 IEEE/CVF Conference on Computer Vision and Pattern
  Recognition (CVPR)}, pages 13376--13386, 2020.

\bibitem[Nguyen et~al.(2017)Nguyen, Hong, Kim, and Park]{nguyen2017person}
Dat~Tien Nguyen, Hyung~Gil Hong, Ki~Wan Kim, and Kang~Ryoung Park.
\newblock Person recognition system based on a combination of body images from
  visible light and thermal cameras.
\newblock \emph{Sensors}, 17\penalty0 (3):\penalty0 605, 2017.

\bibitem[Pan et~al.(2018)Pan, Luo, Shi, and Tang]{pan2018two}
Xingang Pan, Ping Luo, Jianping Shi, and Xiaoou Tang.
\newblock Two at once: Enhancing learning and generalization capacities via
  ibn-net.
\newblock In \emph{Proceedings of the European Conference on Computer Vision
  (ECCV)}, pages 464--479, 2018.

\bibitem[Radenović et~al.(2019)Radenović, Tolias, and Chum]{8382272}
Filip Radenović, Giorgos Tolias, and Ondřej Chum.
\newblock Fine-tuning cnn image retrieval with no human annotation.
\newblock \emph{IEEE Transactions on Pattern Analysis and Machine
  Intelligence}, 41\penalty0 (7):\penalty0 1655--1668, 2019.

\bibitem[Ren et~al.(2021)Ren, He, Liao, Liu, Wang, and Tan]{9711131}
Min Ren, Lingxiao He, Xingyu Liao, Wu Liu, Yunlong Wang, and Tieniu Tan.
\newblock Learning instance-level spatial-temporal patterns for person
  re-identification.
\newblock In \emph{2021 IEEE/CVF International Conference on Computer Vision
  (ICCV)}, pages 14910--14919, 2021.

\bibitem[Selvaraju et~al.(2017)Selvaraju, Cogswell, Das, Vedantam, Parikh, and
  Batra]{selvaraju2017grad}
Ramprasaath~R Selvaraju, Michael Cogswell, Abhishek Das, Ramakrishna Vedantam,
  Devi Parikh, and Dhruv Batra.
\newblock Grad-cam: Visual explanations from deep networks via gradient-based
  localization.
\newblock In \emph{Proceedings of the IEEE international conference on computer
  vision}, pages 618--626, 2017.

\bibitem[Sun et~al.(2018)Sun, Zheng, Yang, Tian, and Wang]{sun2018beyond}
Yifan Sun, Liang Zheng, Yi Yang, Qi Tian, and Shengjin Wang.
\newblock Beyond part models: Person retrieval with refined part pooling (and a
  strong convolutional baseline).
\newblock In \emph{Proceedings of the European conference on computer vision
  (ECCV)}, pages 480--496, 2018.

\bibitem[Wang et~al.(2019{\natexlab{a}})Wang, Zhang, Cheng, Liu, Yang, and
  Hou]{9009814}
Guan'an Wang, Tianzhu Zhang, Jian Cheng, Si Liu, Yang Yang, and Zengguang Hou.
\newblock Rgb-infrared cross-modality person re-identification via joint pixel
  and feature alignment.
\newblock In \emph{2019 IEEE/CVF International Conference on Computer Vision
  (ICCV)}, pages 3622--3631, 2019{\natexlab{a}}.

\bibitem[Wang et~al.(2020)Wang, Zhang, Yang, Cheng, Chang, Liang, and
  Hou]{wang2020cross}
Guan-An Wang, Tianzhu Zhang, Yang Yang, Jian Cheng, Jianlong Chang, Xu Liang,
  and Zeng-Guang Hou.
\newblock Cross-modality paired-images generation for rgb-infrared person
  re-identification.
\newblock In \emph{Proceedings of the AAAI conference on artificial
  intelligence}, pages 12144--12151, 2020.

\bibitem[Wang et~al.(2019{\natexlab{b}})Wang, Wang, Zheng, Chuang, and
  Satoh]{8953262}
Zhixiang Wang, Zheng Wang, Yinqiang Zheng, Yung-Yu Chuang, and Shin'ich Satoh.
\newblock Learning to reduce dual-level discrepancy for infrared-visible person
  re-identification.
\newblock In \emph{2019 IEEE/CVF Conference on Computer Vision and Pattern
  Recognition (CVPR)}, pages 618--626, 2019{\natexlab{b}}.

\bibitem[Wei et~al.(2021)Wei, Yang, Wang, and Gao]{9711501}
Ziyu Wei, Xi Yang, Nannan Wang, and Xinbo Gao.
\newblock Syncretic modality collaborative learning for visible infrared person
  re-identification.
\newblock In \emph{2021 IEEE/CVF International Conference on Computer Vision
  (ICCV)}, pages 225--234, 2021.

\bibitem[Wu et~al.(2017)Wu, Zheng, Yu, Gong, and Lai]{8237837}
Ancong Wu, Wei-Shi Zheng, Hong-Xing Yu, Shaogang Gong, and Jianhuang Lai.
\newblock Rgb-infrared cross-modality person re-identification.
\newblock In \emph{2017 IEEE International Conference on Computer Vision
  (ICCV)}, pages 5390--5399, 2017.

\bibitem[Wu et~al.(2021)Wu, Dai, Chen, Lin, Wu, Huang, Zhong, and Ji]{9577837}
Qiong Wu, Pingyang Dai, Jie Chen, Chia-Wen Lin, Yongjian Wu, Feiyue Huang,
  Bineng Zhong, and Rongrong Ji.
\newblock Discover cross-modality nuances for visible-infrared person
  re-identification.
\newblock In \emph{2021 IEEE/CVF Conference on Computer Vision and Pattern
  Recognition (CVPR)}, pages 4328--4337, 2021.

\bibitem[Yang et~al.(2020)Yang, Zhang, Cheng, Hou, Tiwari, Pandey,
  et~al.]{yang2020cross}
Yang Yang, Tianzhu Zhang, Jian Cheng, Zengguang Hou, Prayag Tiwari, Hari~Mohan
  Pandey, et~al.
\newblock Cross-modality paired-images generation and augmentation for
  rgb-infrared person re-identification.
\newblock \emph{Neural Networks}, 128:\penalty0 294--304, 2020.

\bibitem[Ye et~al.(2020{\natexlab{a}})Ye, Lan, Leng, and Shen]{9107428}
Mang Ye, Xiangyuan Lan, Qingming Leng, and Jianbing Shen.
\newblock Cross-modality person re-identification via modality-aware
  collaborative ensemble learning.
\newblock \emph{IEEE Transactions on Image Processing}, 29:\penalty0
  9387--9399, 2020{\natexlab{a}}.

\bibitem[Ye et~al.(2020{\natexlab{b}})Ye, Shen, J.~Crandall, Shao, and
  Luo]{ye2020dynamic}
Mang Ye, Jianbing Shen, David J.~Crandall, Ling Shao, and Jiebo Luo.
\newblock Dynamic dual-attentive aggregation learning for visible-infrared
  person re-identification.
\newblock In \emph{Computer Vision--ECCV 2020: 16th European Conference,
  Glasgow, UK, August 23--28, 2020, Proceedings, Part XVII 16}, pages 229--247.
  Springer, 2020{\natexlab{b}}.

\bibitem[Ye et~al.(2021{\natexlab{a}})Ye, Ruan, Du, and Shou]{9711494}
Mang Ye, Weijian Ruan, Bo Du, and Mike~Zheng Shou.
\newblock Channel augmented joint learning for visible-infrared recognition.
\newblock In \emph{2021 IEEE/CVF International Conference on Computer Vision
  (ICCV)}, pages 13547--13556, 2021{\natexlab{a}}.

\bibitem[Ye et~al.(2021{\natexlab{b}})Ye, Shen, and Shao]{9115075}
Mang Ye, Jianbing Shen, and Ling Shao.
\newblock Visible-infrared person re-identification via homogeneous augmented
  tri-modal learning.
\newblock \emph{IEEE Transactions on Information Forensics and Security},
  16:\penalty0 728--739, 2021{\natexlab{b}}.

\bibitem[Ye et~al.(2022)Ye, Shen, Lin, Xiang, Shao, and Hoi]{9336268}
Mang Ye, Jianbing Shen, Gaojie Lin, Tao Xiang, Ling Shao, and Steven C.~H. Hoi.
\newblock Deep learning for person re-identification: A survey and outlook.
\newblock \emph{IEEE Transactions on Pattern Analysis and Machine
  Intelligence}, 44\penalty0 (6):\penalty0 2872--2893, 2022.

\bibitem[Yu et~al.(2023)Yu, Cheng, Peng, Liu, and Zhao]{yu2023modality}
Hao Yu, Xu Cheng, Wei Peng, Weihao Liu, and Guoying Zhao.
\newblock Modality unifying network for visible-infrared person
  re-identification.
\newblock In \emph{Proceedings of the IEEE/CVF International Conference on
  Computer Vision}, pages 11185--11195, 2023.

\bibitem[Zhang et~al.(2023)Zhang, Liu, Zhang, and Zhang]{zhang2023style}
Lei Zhang, Zhipu Liu, Wensheng Zhang, and David Zhang.
\newblock Style uncertainty based self-paced meta learning for generalizable
  person re-identification.
\newblock \emph{IEEE Transactions on Image Processing}, 32:\penalty0
  2107--2119, 2023.

\bibitem[Zhang et~al.(2022{\natexlab{a}})Zhang, Lai, Liu, Huang, and
  Han]{9880449}
Qiang Zhang, Changzhou Lai, Jianan Liu, Nianchang Huang, and Jungong Han.
\newblock Fmcnet: Feature-level modality compensation for visible-infrared
  person re-identification.
\newblock In \emph{2022 IEEE/CVF Conference on Computer Vision and Pattern
  Recognition (CVPR)}, pages 7339--7348, 2022{\natexlab{a}}.

\bibitem[Zhang and Wang(2023)]{zhang2023diverse}
Yukang Zhang and Hanzi Wang.
\newblock Diverse embedding expansion network and low-light cross-modality
  benchmark for visible-infrared person re-identification.
\newblock In \emph{Proceedings of the IEEE/CVF Conference on Computer Vision
  and Pattern Recognition}, pages 2153--2162, 2023.

\bibitem[Zhang et~al.(2018)Zhang, Xiang, Hospedales, and Lu]{zhang2018deep}
Ying Zhang, Tao Xiang, Timothy~M Hospedales, and Huchuan Lu.
\newblock Deep mutual learning.
\newblock In \emph{Proceedings of the IEEE conference on computer vision and
  pattern recognition}, pages 4320--4328, 2018.

\bibitem[Zhang et~al.(2022{\natexlab{b}})Zhang, Zhao, Kang, and
  Shen]{zhang2022modality}
Yiyuan Zhang, Sanyuan Zhao, Yuhao Kang, and Jianbing Shen.
\newblock Modality synergy complement learning with cascaded aggregation for
  visible-infrared person re-identification.
\newblock In \emph{Computer Vision--ECCV 2022: 17th European Conference, Tel
  Aviv, Israel, October 23--27, 2022, Proceedings, Part XIV}, pages 462--479.
  Springer, 2022{\natexlab{b}}.

\bibitem[Zhao et~al.(2022)Zhao, Wang, Zhou, Yao, Chen, and El~Saddik]{9745797}
Jiaqi Zhao, Hanzheng Wang, Yong Zhou, Rui Yao, Silin Chen, and Abdulmotaleb
  El~Saddik.
\newblock Spatial-channel enhanced transformer for visible-infrared person
  re-identification.
\newblock \emph{IEEE Transactions on Multimedia}, pages 1--1, 2022.

\bibitem[Zheng et~al.(2022)Zheng, Chen, and Lu]{9935815}
Xiangtao Zheng, Xiumei Chen, and Xiaoqiang Lu.
\newblock Visible-infrared person re-identification via partially interactive
  collaboration.
\newblock \emph{IEEE Transactions on Image Processing}, 31:\penalty0
  6951--6963, 2022.

\bibitem[Zhou et~al.(2019)Zhou, Yang, Cavallaro, and Xiang]{zhou2019omni}
Kaiyang Zhou, Yongxin Yang, Andrea Cavallaro, and Tao Xiang.
\newblock Omni-scale feature learning for person re-identification.
\newblock In \emph{Proceedings of the IEEE/CVF international conference on
  computer vision}, pages 3702--3712, 2019.

\bibitem[Zhu et~al.(2020)Zhu, Yang, Wang, Zhao, Hu, and Tao]{zhu2020hetero}
Yuanxin Zhu, Zhao Yang, Li Wang, Sai Zhao, Xiao Hu, and Dapeng Tao.
\newblock Hetero-center loss for cross-modality person re-identification.
\newblock \emph{Neurocomputing}, 386:\penalty0 97--109, 2020.

\end{thebibliography}
}


\end{document}